# Mesh-based Super-resolution of Multiscale Detonation Flows with Graph Transformers

Shivam Barwey[a], Pinaki Pal[a,*]

[a]*Transportation and Power Systems Division, Argonne National Laboratory, 9700 S. Cass Ave, Lemont, IL 60439, USA*

**Abstract**

Super-resolution flow reconstruction using state-of-the-art data-driven techniques is valuable for a variety of applications, such as subgrid/subfilter closure modeling, accelerating spatiotemporal forecasting, data compression, and serving as an upscaling tool for sparse experimental measurements. In the present work, a first-of-its-kind graph transformer approach is developed for mesh-based super-resolution (SR-GT) of reacting flows. The novel data-driven modeling paradigm leverages a graph-based flow-field representation compatible with complex geometries and non-uniform/unstructured grids. Further, the transformer backbone captures long-range dependencies between different parts of the low-resolution flow-field, identifies important features, and then generates the super-resolved flow-field that preserves those features at a higher resolution. The performance of SR-GT is demonstrated in the context of spectral-element-discretized meshes for a challenging test problem of 2D detonation propagation within a premixed hydrogen-air mixture exhibiting highly complex multiscale reacting flow behavior. The SR-GT framework utilizes a unique element + neighborhood graph representation for the *coarse* input, which is then tokenized before being processed by the transformer component to produce the *fine* output. It is demonstrated that SR-GT provides high super-resolution accuracy for reacting flow-field features and superior performance compared to traditional interpolation-based SR.

*Corresponding author.

## 1. Introduction

High-fidelity computational fluid dynamics (CFD) simulations of turbulent reacting flows play a key role in enhancing our understanding of complex unsteady physico-chemical phenomena prevalent in a wide variety of combustion devices, such as internal combustion engines, gas turbines, scramjets, and detonation engines. However, due to the need to fully resolve highly disparate flow and chemical spatiotemporal scales along with their multiscale interactions, direct numerical simulations (DNS) remain prohibitively expensive for engineering applications. To address these computational bottlenecks, coarse-grid large-eddy simulation (LES) and Reynolds-Averaged Navier-Stokes (RANS) approaches are typically used with closure models to predict unresolved subfilter/subgrid-scale dynamics, such as turbulent/scalar mixing and turbulence-chemistry interaction [1, 2]. Both physics-based and data-driven closure modeling techniques [3, 4] have been explored extensively in the literature. On the other hand, advanced physics-informed machine/deep learning (ML/DL) methods have shown promise for accelerating stiff chemical kinetic computations in reacting flow simulations [5-9].

Super-resolution (SR) methods are another class of techniques that have been investigated in both non-reacting and reacting flow scenarios, with the goal of reconstructing high-resolution (HR) flow-fields from low-resolution (LR) flow-fields [10, 11]. Data-driven SR methods, in particular, have gained a lot of attention recently due to the increasing availability of high-resolution simulation datasets and the capability of such models to achieve promising results in non-canonical flow settings where many purely physics-based SR strategies break down [12]. Notably, a number of DL architectures have been explored, such as multilayer perceptron (MLP) [13], convolution neural network (CNN) [14], generative adversarial network (GAN) [11, 15], graph neural network (GNN) [16], diffusion model [17], and transformer [18]. Bode *et al.* [11] developed a physics-informed enhanced SR GAN (PIERSGAN) framework for LES subfilter modeling, which was trained with unsupervised DL using adversarial and physics-informed losses. The potential of PIERSGAN was demonstrated as a subfilter model for the filtered momentum and scalar equations in reacting LES of Engine Combustion Network (ECN) Spray A case. In a subsequent study, Bode *et al.* [15] further extended the PIERSGAN approach to turbulent premixed combustion. Sofos *et al.* [19] showcased a U-net based neural network architecture for spatiotemporal SR forecasting of two-dimensional (2D) flow-fields featuring shock-wave turbulent boundary layer interaction. Pang *et al.* [20] proposed a Pyramid Res U SR (PSUSR) network and tested it on a practical problem of visualizing turbulent jet flames. Although beneficial, GANs suffer from major issues like training instability and mode collapse. Recently, Barwey *et al.* [16] developed a novel geometric DL framework based on multiscale GNNs for mesh-based SR (SR-GNN) of turbulent flows, which is naturally compatible with complex geometries represented as arbitrary point clouds. The efficacy of SR-GNN was demonstrated for a variety of three-dimensional (3D) flow configurations (Taylor-Green Vortex, Backward-facing Step, Cavity) in the context of spectral-element-discretized meshes, wherein an element-local formulation was employed, leveraging graph representations of mesh neighborhoods of spectral elements. Xu *et al.* [18] proposed a transformer-based DL framework for high-quality SR of turbulent flow-fields.

In light of the above discussion, this work presents a first-of-its-kind graph transformer approach for mesh-based SR (SR-GT) of reacting flows. This novel data-driven modeling paradigm, on one hand, leverages a graph-based flow-field representation compatible with complex geometries and non-uniform/unstructured grids. On the other hand, the transformer backbone captures long-range dependencies between different parts of the LR flow-field, identifies important features, and then generates the super-resolved flow-field that preserves those features at a higher resolution. The performance of SR-GT is demonstrated in the context of spectral-element-discretized meshes for a challenging test problem of 2D detonation propagation within a premixed hydrogen-air mixture exhibiting highly complex multiscale reacting flow behavior.

## 2. Datasets, SR-GT methodology, and key results

### *2.1 Dataset generation and pre-processing*

The SR-GT training and evaluation datasets are generated from a high-fidelity detailed numerical simulation of premixed detonation wave propagating through a stoichiometric hydrogen ($H_2$)-air mixture in a 2D domain at an unburnt static pressure ($P$) of 0.4 bar and static temperature ($T$) of 300 K. The simulation was performed using a high-order Nek5000 discontinuous Galerkin spectral element method (DGSEM) solver [21] developed at Argonne National Laboratory and employed 350,000 non-overlapping spectral elements with uniform spacing of 100 microns (5 elements per induction length) and polynomial order (p) of 3 for each element (resulting in $(3+1)^2 = 16$ non-uniformly arranged Gauss-Lobatto-Legendre (GLL) points within each element for high-order discretization). The top and bottom boundaries employed the slip-wall condition, whereas the left and right boundaries were specified as inflow and outflow, respectively. A 9-species chemical kinetic mechanism from Ó Conaire *et al.* [22] was used for $H_2$/air detailed chemistry. The overall dataset includes 350 instantaneous reacting flow-field snapshots, with uniform temporal spacing of $5e^{-8}$ s, extracted from the quasi-steady detonation propagation phase, out of which the first 300 snapshots comprise the training/validation (70/30%) dataset and the remaining 50 snapshots constitute the test set. The SR goal is to obtain an instantaneous mapping that lifts an LR flow-field (p = 1) to its HR (p = 3) counterpart. To facilitate this, an LR flow-field is generated for each time instant through a masking operation on the corresponding HR flow-field.

It is noted that the SR-GT model (discussed in Sec. 2.2) in the present work leverages a localized reconstruction model with mesh elements cast as individual tokens. In particular, the mesh-based flow-field is converted to a graph representation as follows: a query element is extracted and tokenized alongside its neighbors, with token positions cast as element centroid locations relative to the query, and token dimensions determined by the element GLL points multiplied by the features-per-point. A K-nearest neighbor (KNN) algorithm based on the query element centroid is used to construct a neighborhood graph, with "K" as a hyperparameter providing the query neighborhood size. The elements in this localized neighborhood graph are the input tokens to the transformer. This bakes an inductive bias into the SR-GT model, as it now assumes elements in close proximity to the query element are expected to be responsible for flow super-resolution, eliminating the need to operate on entire flow-fields. This approach of incorporating coarse-element neighborhood has been shown to improve SR accuracy in authors' previous work [16]. A parametric sweep was conducted for different query neighborhood sizes (0, 8, 26, 48) and best performance was achieved for size of 26.

To generate the training/validation dataset, k-means clustering is first applied to each *coarse* (p = 1) flow-field in the training/validation set. It was found that k = 5 and clustering based on $P$, $T$, density ($\rho$), and $H_2O$ mass fraction ($Y_{H2O}$) provided a reasonable segmentation of the detonation flow-field as shown in Fig. 1. Thereafter, cluster-conditioned random sampling is employed, wherein 450 *coarse* query elements are randomly sampled from each cluster, so that different regions of the flow-field are represented in a balanced manner. A *coarse* query element along with its coarse-element neighborhood and the corresponding super-resolved *fine* (p = 3) target query element serves as a single input/output sample.

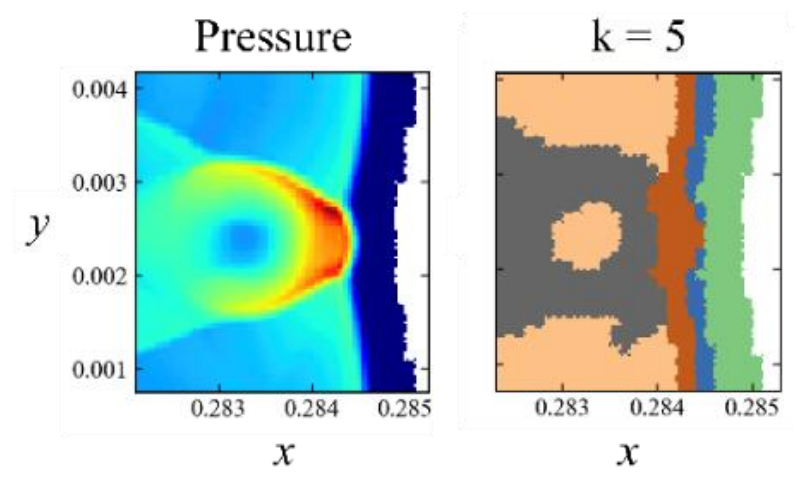

Fig. 1: Zoomed-in view of: (a) pressure field; (b) segmented flow-field based on k-means clustering.

### *2.2 SR-GT model architecture and training approach*

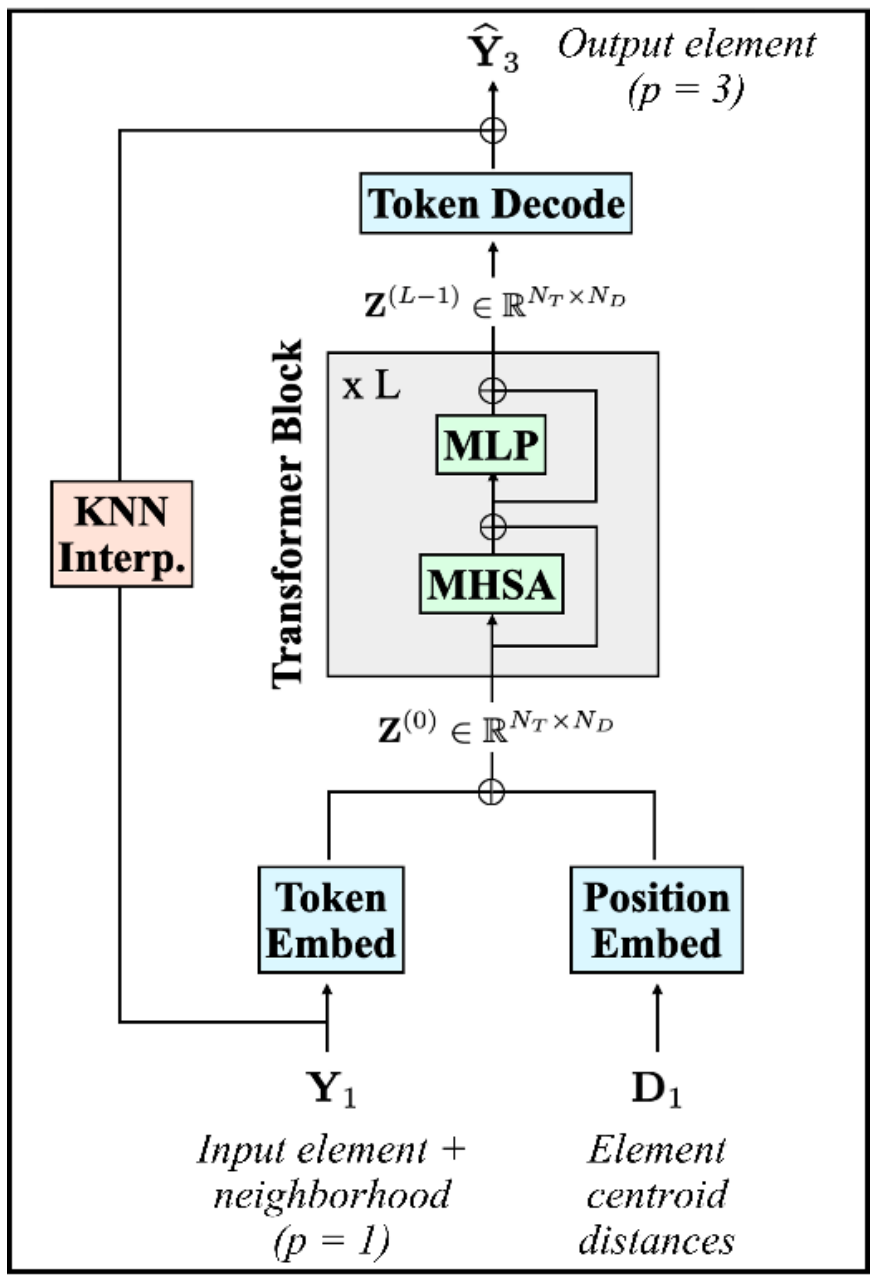

Fig. 2: A schematic of SR-GT model architecture.

A schematic of the SR-GT architecture is depicted in Fig. 2. Evidently, for a query element, SR-GT models a residual which is added to the KNN-interpolated p = 3 flow-field (obtained from the p = 1 flow-field) to get the final super-resolved p = 3 flow-field prediction.

$$\widehat{\mathbf{y}}^0_{3,i} = \widetilde{\mathrm{Interp}(\mathbf{y}^0_{1,i})} + \mathrm{Model}(\widetilde{\mathbf{y}}^0_{1,i}) \tag{1}$$

Where superscript 0 refers to a query element, $i$ refers to the sample number of the query element (total samples = $N_t$), and subscripts 1 and 3 correspond to the polynomial order. The flow-field ($\boldsymbol{y}$) features include velocity components ($u_x$, $u_y$), $P$, $T$, and 9 species mass fractions. The objective loss function is given by the average mean-squared error (MSE) in the predicted *fine* flow-field:

$$\mathcal{L} = \langle \mathrm{MSE}(\widehat{\mathbf{y}}^0_{3,i}, \widetilde{\mathbf{y}}^0_{3,i}) \rangle \tag{2}$$

The mean in the MSE is taken over all the *fine* query element graph nodes and their features. The brackets in Eq. (2) represent an average over all elements in the training (or validation) set. It must be noted that the loss for each query element is computed based on feature-wise scaled flow-fields. Specifically, standard normalization is performed (as a pre-processing step before the SR-GT evaluation) using feature-wise normalization statistics (means and standard deviations) derived from the full neighborhood of the query element.

Since the transformer block(s) receive a sequence of token embeddings as the input, tokenization of the *coarse* query element + neighborhood flow-field is performed as follows:

$$\mathbf{Y}_{p,i} = \begin{pmatrix} \mathbf{y}^0_{p,i} \\ \mathbf{y}^1_{p,i} \\ \vdots \\ \mathbf{y}^{|\mathcal{N}(i)|}_{p,i} \end{pmatrix} \in \mathbb{R}^{N_t \times N_f (p+1)^d} \tag{3}$$

$$\mathbf{y}^0_{p,i} \in \mathbb{R}^{(p+1)^d \times N_f} \tag{4}$$

$$\mathbf{y}^j_{p,i} \in \mathbb{R}^{(p+1)^d \times N_f}, \quad j = 1, \ldots, |\mathcal{N}(i)| \tag{5}$$

Here, $N_f$ refers to the total number of flow-field variables/features (listed earlier), $d = 2$ is the number of spatial dimensions, $\mathbf{y}^0_{p,i}$ denotes the feature set corresponding to all the GLL points in the query element, and Eq. (5) denotes the same for all the neighborhood elements ($N(i) = 26$ as mentioned in Sec. 2.1). In addition to flow-field embedding, embeddings for the relative element positions based on the location of element centroids ($\boldsymbol{r}^j_i$) are also incorporated as follows:

$$\mathbf{D}_{p,i} = \begin{pmatrix} \mathbf{u}^0_{p,i}, d^0_{p,i} \\ \mathbf{u}^1_{p,i}, d^1_{p,i} \\ \vdots \\ \mathbf{u}^{|\mathcal{N}(i)|}_{p,i}, d^{|\mathcal{N}(i)|}_{p,i} \end{pmatrix} \in \mathbb{R}^{N_t \times d+1} \tag{6}$$

$$\mathbf{u}^j_{p,i} = \frac{\bar{\mathbf{r}}^j_i - \bar{\mathbf{r}}^0_i}{\|\bar{\mathbf{r}}^j_i - \bar{\mathbf{r}}^0_i\|_2}, \quad d^j_{p,i} = \|\bar{\mathbf{r}}^j_i - \bar{\mathbf{r}}^0_i\|_2 \tag{7}$$

The transformer uses constant latent vector size $N_D$ (= 256) through all of its layers. $L$ (= 8) transformer blocks are employed. A key feature of each transformer block is the multi-head self-attention (MHSA) module [23], in which multiple self-attention operations, called "heads", are run in parallel and their concatenated outputs are projected. Head dimensionality of 64 was used, resulting in 4 self-attention heads. In each transformer block, the MHSA is followed by an MLP. SR-GT training strategy involves mini-batching with a batch size of 64, AdamW optimizer, dropout with a probability of 0.1, gradient clipping, and early stopping (based on the validation loss decay). In addition, a custom learning rate scheduler is employed, similar to Lam *et al.* [24], consisting of 3 phases: in phase 1, the learning rate is linearly increased from zero to $1e^{-4}$ (warm-up); in phase 2, the learning rate is cosine-decayed to a final learning rate of $1e^{-7}$; in phase 3, the final learning rate is held constant. The overall data-driven modeling framework was implemented using a combination of PyTorch [25] and PyTorch Geometric [26] libraries. Model was trained using 4 Nvidia A100 GPUs using one node of the Polaris supercomputer at the Argonne Leadership Computing Facility (ALCF).

### *2.3. Results and discussion*

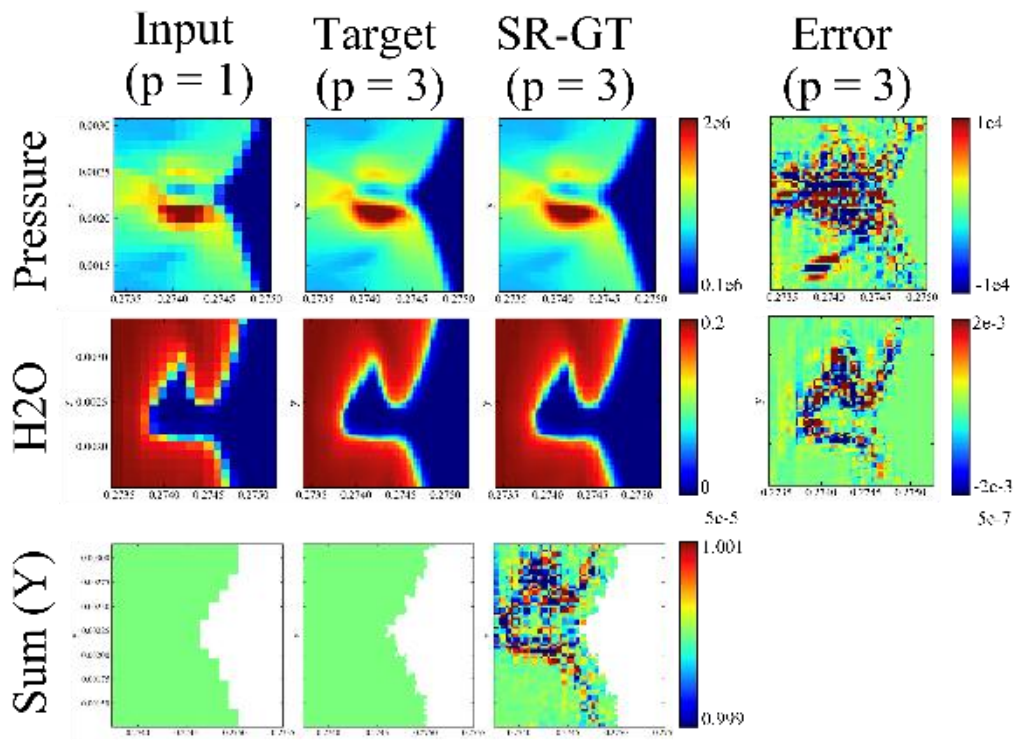

Fig. 3: Zoomed-in view of the input, target, and SR-GT-predicted spatial distributions of static pressure, $H_2O$ mass fraction and sum of mass fractions for a test snapshot. The SR-GT prediction errors for static pressure and $H_2O$ mass fraction are also visualized.

Figure 3 shows the performance of SR-GT on a test snapshot with respect to super-resolution of static pressure and $H_2O$ mass fraction fields. In addition, the degree to which the super-resolved flow-field adheres to total mass conservation is also depicted. It can be clearly seen that SR-GT performs very well with errors within 1% approximately. Moreover, even though mass conservation constraint was not explicitly imposed via either model architecture or training approach, SR-GT predicted flow-field still shows negligible violation of <= 0.1%. The super-resolution quality of SR-GT is further assessed against KNN interpolation. Figure 4 illustrates the spatial error distributions for static temperature fields predicted by SR-GT and KNN interpolation for a test snapshot; only a portion of the domain capturing the detonation front and 30 induction lengths behind the detonation front is shown. Evidently, SR-GT leads to much lower errors than KNN interpolation very close to the detonation front. Further, the relatively high-error regions for SR-GT are primarily concentrated close to the detonation front, whereas KNN-interpolated field shows much higher errors throughout the flow-field. This indicates the superiority of SR-GT compared to traditional interpolation-based SR schemes.

It is noted that although SR-GT is demonstrated here for a regular mesh configuration, it is naturally compatible with complex 2D/3D geometries as unstructured flow-fields can be readily represented as graphs, which will be explored in future work. Other future research directions are closure modeling and spatiotemporal SR forecasting [18] demonstrations for turbulent reacting flows, incorporation of physical laws into the training framework, and showcasing transfer learning capabilities.

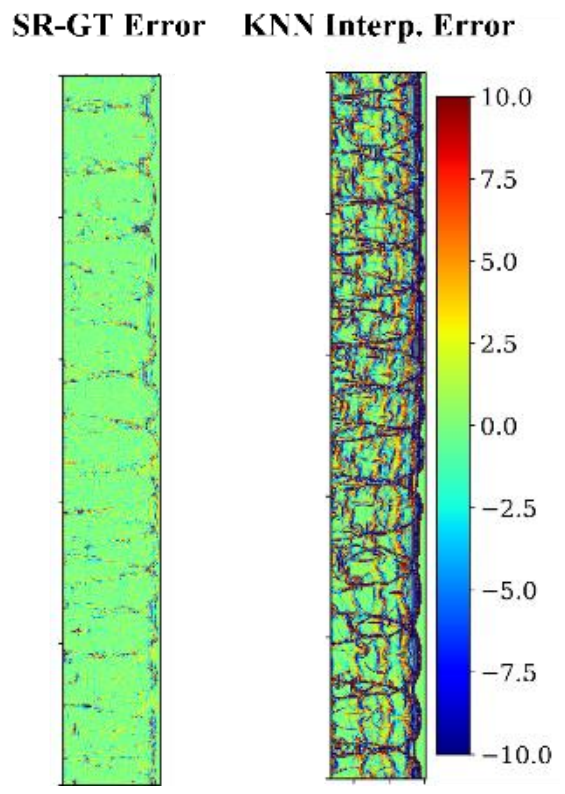

Fig. 4: Visualization of errors (legend in Kelvin) in the super-resolved temperature fields predicted by SR-GT and KNN interpolation for a test snapshot.

## Acknowledgements

The manuscript has been created by UChicago Argonne, LLC, Operator of Argonne National Laboratory (Argonne). The U.S. Government retains for itself, and others acting on its behalf, a paid-up nonexclusive, irrevocable world-wide license in said article to reproduce, prepare derivative works, distribute copies to the public, and perform publicly and display publicly, by or on behalf of the Government. The authors acknowledge Laboratory-Directed Research and Development (LDRD) funding support from Argonne's Advanced Energy Technologies (AET) directorate. Dr. Benjamin Keeton is acknowledged for sharing the 2D detonation simulation datasets. This research used resources of the ALCF.